\documentclass{article}

\pdfoutput=1
\usepackage{arxiv}

\usepackage[utf8]{inputenc} % allow utf-8 input
\usepackage[T1]{fontenc}    % use 8-bit T1 fonts
\usepackage{hyperref}       % hyperlinks
\usepackage{url}            % simple URL typesetting
\usepackage{booktabs}       % professional-quality tables
\usepackage{amsfonts}       % blackboard math symbols
\usepackage{amsmath}
\newcommand\numberthis{\addtocounter{equation}{1}\tag{\theequation}}

\usepackage{nicefrac}       % compact symbols for 1/2, etc.
\usepackage{microtype}      % microtypography
\usepackage{lipsum}
\usepackage{graphicx}
\usepackage[natbibapa]{apacite}
\graphicspath{ {./images/} }

\usepackage{graphicx}
\usepackage[ruled,vlined,noend]{algorithm2e}
\usepackage{subcaption}
\usepackage{float}
\usepackage{mwe}

\title{Investigating the Scalability and Biological Plausibility of the Activation Relaxation Algorithm}

\author{
 Beren Millidge \\
  School of Informatics\\
  University of Edinburgh\\
  \texttt{beren@millidge.name}
   \And
 Alexander Tschantz \\
  Sackler Center for Consciousness Science\\
  School of Engineering and Informatics\\
  University   Sussex \\
  \texttt{tschantz.alec@gmail.com} \\
  \And
  Anil K Seth \\
  Sackler Center for Consciousness Science\\
   Evolutionary and Adaptive Systems Research Group\\
  School of Engineering and Informatics\\
  University of Sussex\\
  \texttt{A.K.Seth@sussex.ac.uk} \\
  \And
 Christopher L Buckley \\
  Evolutionary and Adaptive Systems Research Group\\
  School of Engineering and Informatics\\
  University of Sussex\\
  \texttt{C.L.Buckley@sussex.ac.uk} 
}

\begin{document}
\maketitle

\begin{abstract}
    The recently proposed Activation Relaxation (AR) algorithm provides a simple and robust approach for approximating the backpropagation of error algorithm using only local learning rules. Unlike competing schemes, it converges to the exact backpropagation gradients, and utilises only a single type of computational unit and a single backwards relaxation phase. We have previously shown that the algorithm can be further simplified and made more biologically plausible by (i) introducing a learnable set of backwards weights, which overcomes the weight-transport problem, and (ii) avoiding the computation of nonlinear derivatives at each neuron. However, tthe efficacy of these simplifications has, so far, only been tested on simple multi-layer-perceptron (MLP) networks. Here, we show that these simplifications still maintain performance using more complex CNN architectures and challenging datasets, which have proven difficult for other biologically-plausible schemes to scale to. We also investigate whether another biologically implausible assumption of the original AR algorithm -- the \emph{frozen feedforward pass} -- can be relaxed without damaging performance.

\end{abstract}

The backpropagation of error algorithm (backprop) has been the engine driving the successes of modern machine learning with deep neural networks. Backprop solves the credit assignment problem, which is crucial for effectively training such large and distributed networks. The credit assignment problem concerns how to correctly distribute credit for a global outcome to each of the many parameters of the network. Backprop solves this problem by exploiting the chain rule of calculus to recursively compute derivatives. The brain, as a distributed neural network comprising trillions of synapses, also faces a formidable credit assignment problem. Since backprop is the optimal way to solve this problem \citep{baldi2016theory}, one might suspect that the brain has evolved to implement a backprop-like mechanism. However, a canonical implementation of backprop is held to be biologically implausible \citep{crick1989recent,lillicrap2020backpropagation}, due to entailing update rules that require information about the global state of the network to be present at each synapse, as well as requiring symmetry of the forward weights, used to compute predictions, and the backwards weights used to send back gradients, which is known as the `weight transport problem'.

Recently, a variety of learning schemes have attempted to overcome this biological implausibility by using only local learning rules \citep{lillicrap2016random,nokland2016direct,scellier2017equilibrium,sacramento2018dendritic}. These schemes broadly fit into two categories. First, `target propagation' schemes which 
sequentially backpropagate targets which can be computed locally, and then rely on a layerwise minimization of the difference between the activity and the local target \citep{lee2015difference,ororbia2019biologically}. %The exemplar of these schemes is targetprop in which the target is computed as the inverse of the optimal activation at the output.
These schemes typically do not exactly approximate the backprop gradients, and it has recently been shown they can be understood in terms of second order optimization \citep{meulemans2020theoretical}. Secondly, `recurrent schemes' exploit recurrent dynamics to enable information about the global output or target to `leak' back through the system via local rules over the course of multiple iterations. Such schemes, which include predictive coding \citep{whittington2017approximation,millidge2020predictive} and equilibrium-prop \citep{scellier2018extending,scellier2018generalization}, have been shown to exactly approximate backprop using only local learning rules. However, they typically require multiple dynamical iterations to converge. In a companion paper, we introduced a novel recurrent scheme -- Activation Relaxation (AR) -- which eliminates much of the complexity of previous schemes while preserving their asymptotic performance \citep{millidge2020activation}. AR operates in two phases. The first `activation' phase is a standard feedforward sweep through the network. In the second `relaxation' phase, the activities of each unit are dynamically updated according to a simple local learning rule. At convergence, the weights of the network are updated using the activities computed in the relaxation phase. In the companion paper, we show that AR converges rapidly and robustly to the exact backprop gradients and can be used to train deep neural networks with equivalent performance to backprop. Moreover, unlike predictive coding, AR only utilizes one type of computational unit. Unlike equilibrium-prop, AR does not require two separate backwards phases (a 'free phase' and a 'clamped phase'), nor the storage of information between backwards phases. Moreover, we described two simplifications of the algorithm to enhance its biological plausibility -- removing the symmetry between backwards and forwards weights through a set of initially random learnable weights, and dropping the nonlinear derivative terms, which might be difficult for biological neurons to compute, and showed that these simplifications did not harm performance in simple MLP models. 

However, the previous paper left open two questions about the effectiveness and biological plausibility of the AR algorithm which we investigate empirically here. The first concerns an important assumption made in the original algorithm, which we call the 
\emph{frozen forward pass} assumption. The original AR algorithm requires that the values of the forward pass be stored and used repeatedly in the relaxation phase updates. Specifically, the original value of the feedforward pass activation and the derivative of the activation function with respect to the feedforward pass activation need to be stored.
Although local, this requirement poses a problem of biological plausibility since it is not clear how these quantities could be maintained by the neuron and utilised in the learning rule during the relaxation phase. In this paper, we empirically investigate the extent to which this assumption can be relaxed. We show that we can successfully relax the requirement to store the nonlinear derivative of the feedforward pass value, but we \emph{cannot} relax the assumption for the activity value in the weight updates. This may limit the applicability of the algorithm, as is, in real neurons, or else require that neurons possess some means of storing this information throughout the relaxation phase.

The second question concerns the scalability of the algorithm and especially the simplifications introduced in the second part of the companion paper. Although the convergence to backprop of the AR scheme is shown on an arbitrary computation graph (see Appendix A), and thus any machine learning architecture of any scale, the efficacy of the simplifications proposed to the algorithm were only tested on a simple MLP architecture on the MNIST and Fashion-MNIST datasets. Here, we show that these simplifications remain applicable on larger scale CNN architectures and on more challenging datasets such as CIFAR10 and CIFAR100, which is notable since scaling up to CNNs has historically been challenging for other similar schemes \citep{bartunov2018assessing,launay2019principled}.
\section{Methods}
The difficulty of neurally implementing backprop is computing the $\frac{\partial L}{\partial x^l}$ term, where $x^l$ denotes the vector of activations of a given layer $l$ of a neural network, and $L$ is the global loss function. The key idea of AR is to directly approximate this quantity by defining a dynamical system which converges to it over the course of a relaxation phase. A simple system which converges to this quantity takes the form of a leaky integrator system, which requires only local quantities and is biologically plausible
\begin{align*}
    \frac{dx^l}{dt} = -x^l + \frac{\partial L}{\partial x^l}
\end{align*}
It can be seen that, at equilibrium, this system converges to the BP gradients,
\begin{align*}
    \frac{dx^l}{dt} = 0 \implies {x^l}^* = \frac{\partial L}{\partial x^l}
\end{align*}
Using the chain rule, we can write this dynamical system as:
\begin{align*}
    \frac{dx^l}{dt} &= -x^l + \frac{\partial L}{\partial x^{l+1}}\frac{\partial x^{l+1}}{\partial x^l} \Bigr|_{x^l=\bar{x}^l} 
     = -x^l + {x^{l+1}}^*\frac{\partial x^{l+1}}{\partial x^l} \Bigr|_{x^l=\bar{x}^l} \\
     &\approx -x^l + {x^{l+1}}\frac{\partial f(W^l \bar{x}^l)}{\partial x^l} {W^l}^T \numberthis
\end{align*}
where $\bar{x}$ is the value of $x$ computed in the feedforward pass. In the second line, we make the crucial approximation $x^* \approx x$ which allows all layers to be updated in parallel without waiting for the layers above to converge to equilibrium. Empirically, we find that this assumption does not prevent convergence to the correct backprop gradients. The second assumption, which we call the `frozen feedforward pass' assumption, is that the derivative $\frac{\partial x^{l+1}}{\partial x^l}$ be computed at $x = \bar{x}$. The weight update equation 
\begin{align*}
    \frac{\partial L}{\partial W^l} = \frac{\partial L}{\partial x^{l+1}}\frac{\partial x^{l+1}}{\partial W^l} \Bigr|_{x^{l+1}=\bar{x}^{l+1}}  = {x^{l+1}}^* \bar{x}^T {\frac{\partial f(W^l \bar{x}^l)}{\partial \bar{x}}}^T \numberthis
\end{align*}
requires both the value of the activity and the nonlinear derivative to be fixed at their feedforward pass values. We note two additional implausibilities. The first is the $W^T$ term in Equation 1 which furnishes the weight transport problem, and the $\frac{\partial f}{\partial x}$ term which results in the backwards nonlinearity problem. We show in the companion paper that the first term can be overcome by replacing the backwards weights $W^T$ with a set of random backwards weights $\psi$, which are learnt with the learning rule
\begin{align*}
   \frac{d \psi^l}{dt} = {\frac{\partial f(W^l \bar{x}^l)}{\partial \bar{x}}} \bar{x} {{x^{l+1}}^*}^T \numberthis
\end{align*}
and we show that the second problem of backwards nonlinearities can be solved by simply dropping the nonlinearity from the Equations 1 and 2, which gives the simple and highly plausible update rule,
\begin{align*}
   \frac{dx^l}{dt} \approx -x^l + {x^{l+1}} \psi^l \numberthis
\end{align*} 
%However, in the companion paper these simplifications were only tested on simple MLPs on the relatively easy MNIST and Fashion-MNIST datasets. Here we investigate whether these simplifications will scale to more challenging CNN architectures and harder datasets. \vspace{-0.4cm}
\section{Results}
First, we investigate whether the frozen feedforward pass assumption can be relaxed. We train a 4-layer MLP network identical to the one used in the companion paper on the MNIST and Fashion-MNIST (see Appendix B for details) datasets under three different conditions. We evaluate whether the nonlinear derivative term can be unfrozen so that it uses the current value of the activity in a.) the relaxation update (Equation 1), b.) in the weight update equation (Equation 2), and c.) we investigate whether the activation value itself can be replaced in the weight update equation.
\begin{figure}[htb]
\centering
  \begin{subfigure}[b]{0.4\linewidth}
    \centering
    \includegraphics[width=0.75\linewidth]{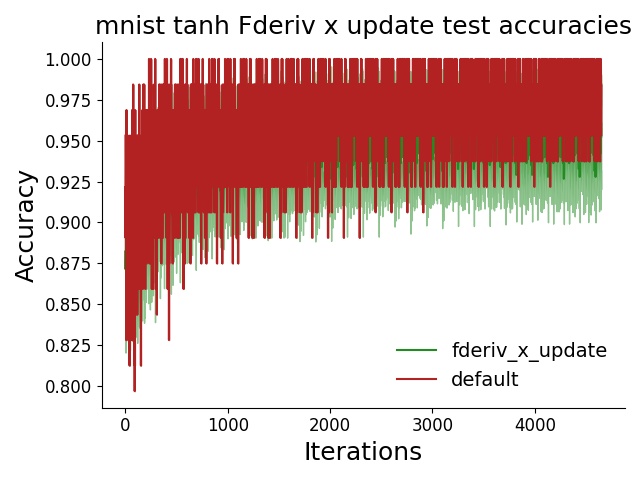} 
    \caption{MNIST nonlinear derivative relaxation update} 
    \vspace{4ex}
  \end{subfigure}%% 
  \begin{subfigure}[b]{0.4\linewidth}
    \centering
    \includegraphics[width=0.75\linewidth]{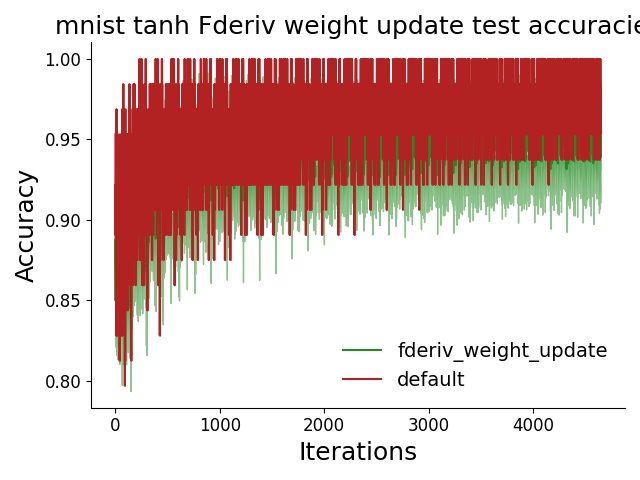} 
    \caption{MNIST nonlinear derivative weight update} 
    \vspace{4ex}
  \end{subfigure} 
  \begin{subfigure}[b]{0.4\linewidth}
    \centering
    \includegraphics[width=0.75\linewidth]{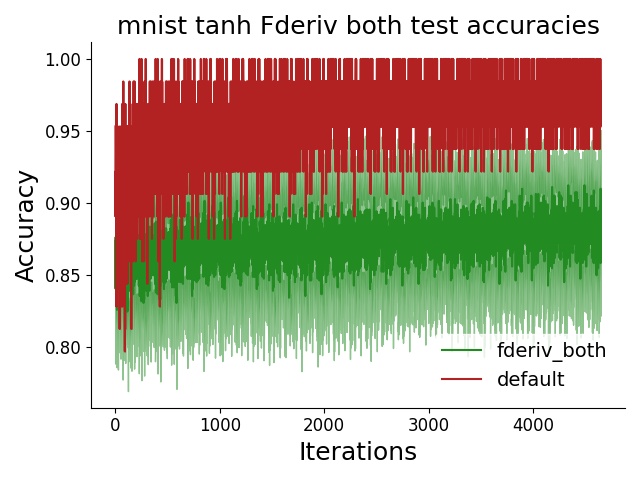} 
    \caption{MNIST both nonlinear derivative} 
  \end{subfigure}%%
  \begin{subfigure}[b]{0.4\linewidth}
    \centering
    \includegraphics[width=0.75\linewidth]{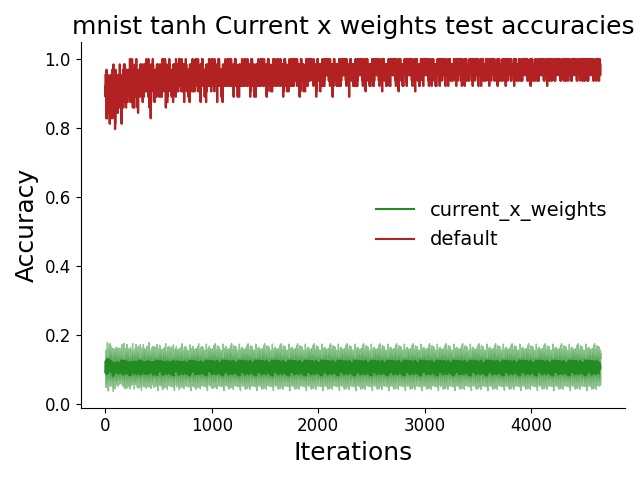} 
    \caption{MNIST current x weight update} 
  \end{subfigure} 
  \caption{Assessing whether the frozen feedforward pass assumption can be relaxed. We show the resulting performance (test accuracy) against baseline of relaxing this assumption on the MNIST dataset. See Appendix B for the Fashion-MNIST results. All results averaged over 10 seeds.} 
\end{figure} 
In Figure 1, we see that the frozen feedforward pass assumption can be relaxed in the case of the nonlinear derivatives for both the AR update and the weight update equation. However, relaxing it in the case of the weight update equation destroys performance. This means that ultimately, a direct implementation of AR in biological circuitry would require neurons to store the feedforward pass value of their own activations. Secondly, we investigate the scalability of the AR algorithm and the simplifications proposed in the companion paper -- learnable backwards weights and dropping nonlinear derivatives -- by testing them on a CNN on more challenging datasets (SVHN, CIFAR, and CIFAR100; results for CIFAR100 and SVHN are presented in Appendix C). The extension to CNNs is important because other biologically plausible schemes such as feedback alignment \citep{lillicrap2016random}, and directed feedback alignment \citep{nokland2016direct}, have been shown to struggle with the CNN architectures \citep{launay2019principled}. We tested learning the feedback weights of both the convolutional layers and the fully connected layers separately \footnote{All code to reproduce the experiments and the figures can be found online at https://github.com/BerenMillidge/Dynamical-Activation-Relaxation}.
 
 \begin{figure}[htb]
    \centering % <-- added
\begin{subfigure}{0.3\textwidth}
  \includegraphics[width=\linewidth]{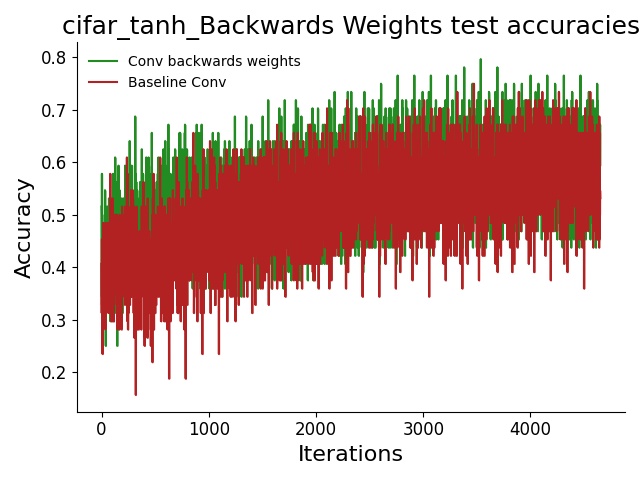}
  \caption{Conv backwards weights}
\end{subfigure}\hfil % <-- added
\begin{subfigure}{0.3\textwidth}
  \includegraphics[width=\linewidth]{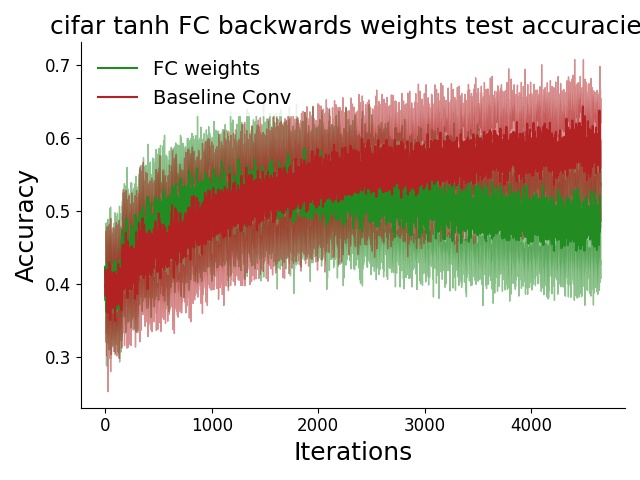}
  \caption{FC backwards weights}
\end{subfigure}\hfil % <-- added
\begin{subfigure}{0.3\textwidth}
  \includegraphics[width=\linewidth]{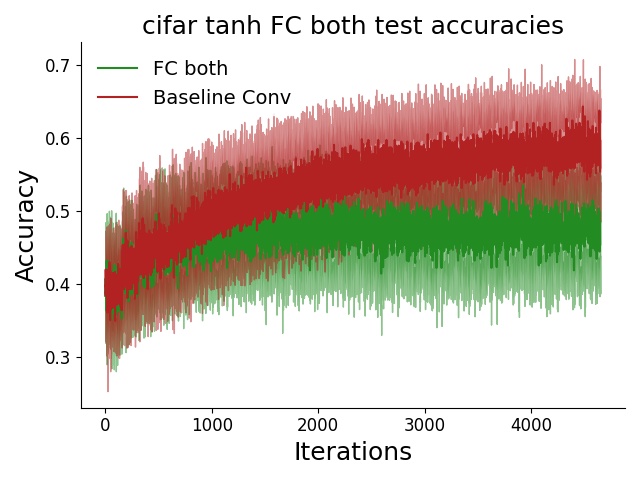}
  \caption{Both backwards weights}
\end{subfigure}

\medskip
\begin{subfigure}{0.3\textwidth}
  \includegraphics[width=\linewidth]{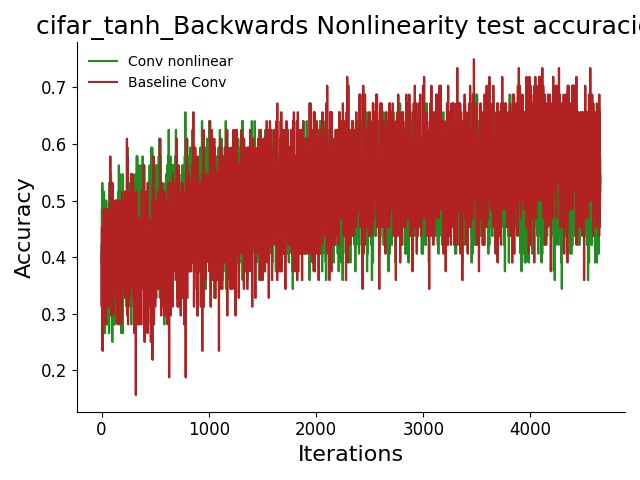}
  \caption{Conv no nonlinear derivative}
\end{subfigure}\hfil % <-- added
\begin{subfigure}{0.3\textwidth}
  \includegraphics[width=\linewidth]{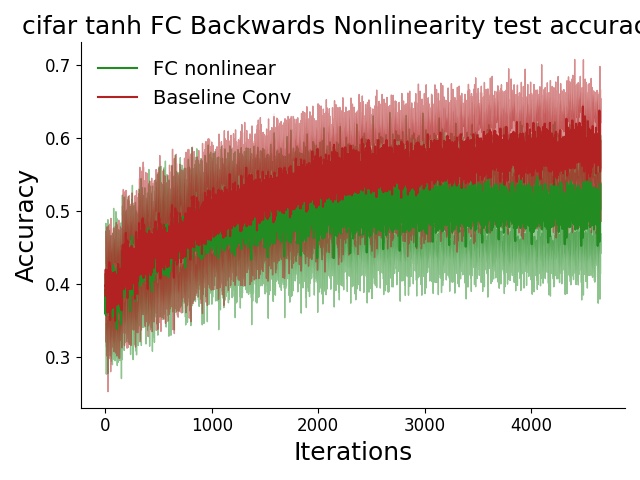}
  \caption{FC no nonlinear derivative}
\end{subfigure}\hfil % <-- added
\begin{subfigure}{0.3\textwidth}
  \includegraphics[width=\linewidth]{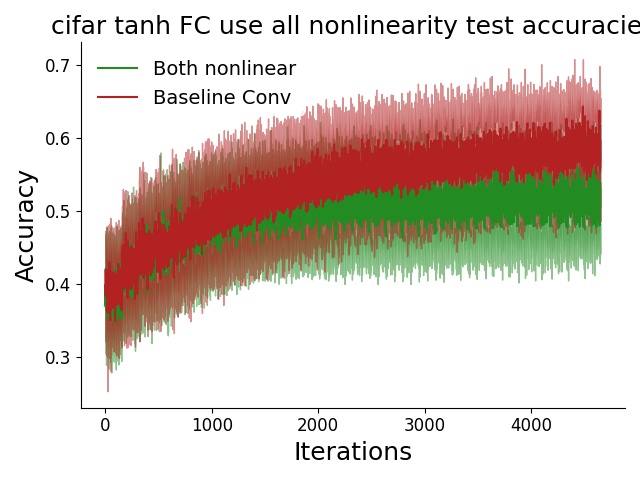}
  \caption{Both no nonlinear derivative}
\end{subfigure}
\caption{Performance (test accuracy), averaged over 10 seeds, on CIFAR10 demonstrating the scalability of the learnable backwards weights and dropping the nonlinear derivatives in a CNN architecture, compared to baseline AR without simplifications. Performance is equivalent throughout.} 
\end{figure} 
\section{Discussion}
In this paper, we empirically investigated the scalability and biological plausibility of AR. Firstly, we studied the degree to which the assumption of the frozen feedforward pass can be relaxed. We showed that it can be relaxed on the backwards nonlinearities in both the relaxation update equation (Equation 1) and in the weight update equation (Equation 2). However, we found that the assumption \emph{cannot} be relaxed for the activity values in the weight update equation. To do so completely destroys performance. Secondly, we investigated whether the simplifications introduced in the companion paper -- using learnable backwards weights to address the weight-transport problem, and dropping the nonlinear derivatives -- are scalable to a more complex CNN architectures. We showed that these simplifications did indeed scale, both alone and in combination, to CNN architectures and more challenging datasets. These results are highly promising for future studies aimed at investigating the potential of the AR algorithm to support biologically plausible learning in complex architectures.

Our first set of results showing that the frozen feedforward pass assumptions can be relaxed on the nonlinear derivative terms but not the activities themselves are mixed. While it is important and interesting that they can be relaxed on the nonlinear derivatives, this is not necessarily surprising given that the nonlinear derivatives can be dropped entirely. More importantly, the fact that to maintain performance the activities in the weight updates \emph{must} be fixed at their feedforward pass values raises a potential hurdle to the biological plausibility of AR, since this value must be stored throughout the relaxation phase such that it can be utilized at the synapse for the weight update. However, this storage may be potentially possible through the use of relatively short-lived synaptic traces, for which there is a fair amount of neurobiological evidence \citep{bellec2020solution}, as long as the relaxation phase is relatively short. Alternatively, using more detailed multicompartment neuron models, the information may be stored in different neural compartments until it is used. An alternative possibility is that the original activations may be maintained in a separate population of neurons during the backwards relaxation phase and the retrieval of this information during the synaptic weight updates could be coordinated by the regular oscillatory phase activity observed in the brain \citep{buzsaki2006rhythms}.

\section{Acknowledgements}

BM is supported by an EPSRC funded PhD Studentship. AT is funded by a PhD studentship from the Dr. Mortimer and Theresa Sackler Foundation and the School of Engineering and Informatics at the University of Sussex. CLB is supported by BBRSC grant number BB/P022197/1 and by Joint Research with the National Institutes of Natural Sciences (NINS), Japan, program No. 01112005.  AT and AKSare grateful to the Dr. Mortimer and Theresa Sackler Foundation, which supports the Sackler Centrefor Consciousness Science.   AKS is additionally grateful to the Canadian Institute for AdvancedResearch (Azrieli Programme on Brain, Mind, and Consciousness).

\bibliography{references}

\section*{Appendix A: The AR algorithm}

In this appendix we present the detailed pseudocode of the AR algorithm and show how it can be extended to arbitrary computation graphs. The AR algorithm operates in two phases. First there is a feedforward pass, as in the inference phase of a standard neural network, which computes the network's estimate of each activity. Then there is a relaxation phase the activities of all layers are simultaneously updated with Equation 1. Upon convergence of the activities, the weights are updated with Equation 2.

\begin{algorithm}[H]
%\DontPrintSemicolon
\SetAlgoLined
\KwData{Dataset $\mathcal{D} = \{\mathbf{X},\mathbf{T}\}$, parameters $\Theta = \{W^0 \dots W^L\}$, inference learning rate $\eta_x$, weight learning rate $\eta_\theta$.} 
\tcc{Iterate over dataset}
\For{$(x^0, t \in \mathcal{D})$}{
\tcc{Initial feedforward sweep} 
\For{$(x^l,W^l)$ for each layer}{
$x^{l+1} = f(W^l, x^l)$ 
}
\tcc{Begin backwards relaxation}
\While{not converged}{
\tcc{Compute final output error}
$\epsilon^L = T - x^L $ \\ 
$ dx^L= -x^L +  \epsilon^L \frac{\partial \epsilon^L}{\partial x^L} $ \\
\For{$x^l, W^l, x^{l+1}$ for each layer}{
\tcc{Activation update} 
 $dx^l = -x^l +  x^{l+1} \frac{\partial x^{l+1} }{\partial x^l}$ \\
${x^l}^{t+1} \leftarrow {x^l}^t + \eta_x dx^l $
}
}
\tcc{Update weights at equilibrium}
\For{$W^l \in \{W^0 \dots W^L \}$}{
${W^l}^{t+1} \leftarrow {W^l}^t + \eta_\theta x^l \frac{\partial x^l}{\partial W^l}$ 
}
}
\caption{Activation Relaxation}
\end{algorithm} 

Next we show that the AR algorithm can be extended to converge to the backprop gradients not just in MLP networks composed of hierarchical layers but on arbitrary \emph{computation graphs}. This extension enables AR to be applied to essentially all modern machine learning architectures. A computation graph represents a complex function (such as the forward pass of a complex NN architecture like a transformer \citep{vaswani2017attention}, or an LSTM \citep{hochreiter1997long}) as a graph of simpler functions. Each function corresponds to a vertex of the graph, while there is a directed edge between two vertices if the parent vertex is an argument to the function represented by the child vertex. Because we only study finite feedforward architectures (and since it is assumed finite we can `unroll' any recurrent network into a long feedforward graph), we can represent any machine learning architecture as a directed acyclic graph (DAG). Automatic Differentiation (AD) techniques, which are at the heart of modern machine learning \citep{griewank1989automatic,van2018automatic,margossian2019review}, can then be used to compute gradients with respect to the parameters of any almost arbitrarily complex architecture automatically. This allows machine-learning practitioners to derive models which encode complex inductive biases about the structure of the problem domain, without having to be manually derive the expressions for the derivatives required to train the models. Here, we show that the AR algorithm can also be used to compute these derivatives along arbitrary DAGs, using only local information in the dynamics,and requiring only the knowledge of the inter-layer derivatives. 

Core to AD is the multivariate chain-rule of calculus. Given a node $x^l$ on a DAG, the derivative with respect to some final output of the graph can be computed recursively with the relation
\begin{align}
    \frac{\partial L}{\partial x^i} = \sum_{x^j \in Chi(x^j)} \frac{\partial L}{\partial x^j} \frac{\partial x^j}{\partial x^i}
\end{align}

Where $Chi(x^i)$ represents all the nodes which are children of $x^i$. In effect, this recursive rule states that the derivative with respect to the loss of a point is equal to the sum of the derivatives coming from all paths from that node to the output. For AR to apply to an arbitrary computation graph, all we require is that the activation update in the relaxation phase (Equation 1) be replaced with the sum of the activities from all it's child nodes.
\begin{align*}
    \frac{dx^i}{dt} &= -x^i + \sum_{x^j \in Chi(x^j)} \frac{\partial L}{\partial x^j} \\
    &= -x^i + \sum_{x^j \in Chi(x^j)} {x^j}^* \frac{\partial x^j}{\partial x^i}  \Bigr|_{x^l=\bar{x}^l} \numberthis \\
    &\approx -x^i + \sum_{x^j \in Chi(x^j)} {x^j} \frac{\partial x^j}{\partial x^i}  \Bigr|_{x^l=\bar{x}^l} \numberthis
\end{align*}

The equilibrium of this dynamic will thus tend towards the correct backprop gradients, and since the relationship between the gradient of the parent and the gradients of the children satisifes the same recursive relationship as the multivariable chain rule (Equation 5) then each node of the graph, and thus the layer as a whole will converge during the relaxation phase to the exact backprop gradients throughout the computation graph.

\section*{Appendix B: MLP architecture and FashionMNIST results}

For these experiments, we used a simple MLP model identical to that used in the companion paper. Specifically, we tested a four layer MLP where each layer consisted of 300,300,100, and 10 neurons respectively. All layers had a hyperbolic tangent activation function except for the final layer which was linear. For the relaxation phase in all experiments throughout this paper we used a learning rate of 0.1. We used a minibatch size of 64. We ran the AR update rule (Equation 2) for 100 iterations which we found empirically was sufficient for convergence. After the relaxation phase had completed, the weights were updated with a learning rate of $0.0005$. The network was trained to predict the correct one-hot label using a mean-square-error loss function. 

The two datasets used were MNIST and Fashion-MNIST \citep{xiao2017online}. The MNIST dataset consists of 60000 training and 10000 test 28x28 grayscale images of handwritten digits. This task is generally considered relatively easy to solve and the state of the art for simple MLP models is approximately 98\%, which we obtain with our network. The Fashion-MNIST dataset is desigend to be a `drop-in' replacement for MNIST, but substantially more challenging to solve. The Fashion-MNIST dataset consists of grayscale 28x28 images of clothing items which must be classified into one of ten classes. All results presented were the average of 10 random seeds. We also plot the standard error as error bars.

Here we show the results for the MLP scheme assessing whether the frozen feedforward pass assumption can be relaxed on the more challenging Fashion-MNIST dataset. Importantly, performance is almost as unhindered compared to the baseline even on this dataset. Trying to relax the frozen feedforward pass assumption for the activity term in the weight update completely destroys performance as on the MNIST dataset.

\begin{figure}[ht] 
  \begin{subfigure}[b]{0.5\linewidth}
    \centering
    \includegraphics[width=0.75\linewidth]{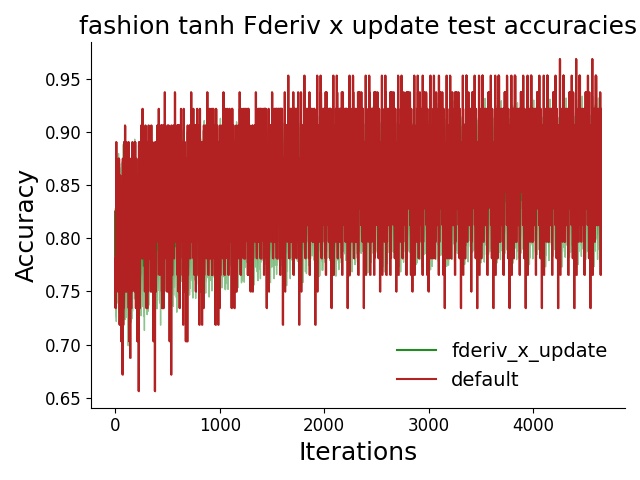} 
    \caption{Fashion nonlinear derivative relaxation update} 
    \vspace{4ex}
  \end{subfigure}%% 
  \begin{subfigure}[b]{0.5\linewidth}
    \centering
    \includegraphics[width=0.75\linewidth]{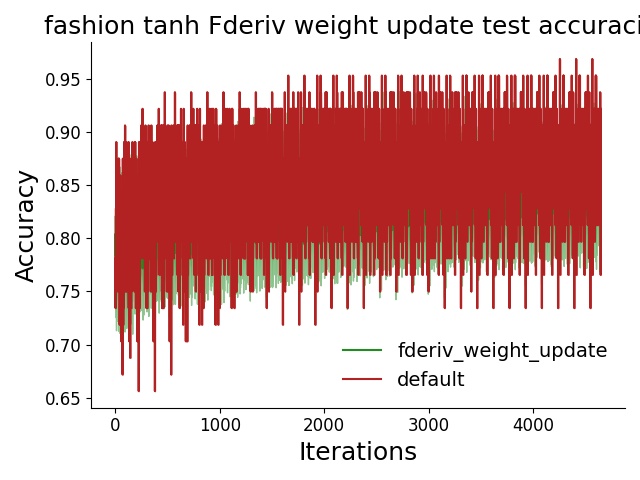} 
    \caption{Fashion nonlinear derivative weight update} 
    \vspace{4ex}
  \end{subfigure} 
  \begin{subfigure}[b]{0.5\linewidth}
    \centering
    \includegraphics[width=0.75\linewidth]{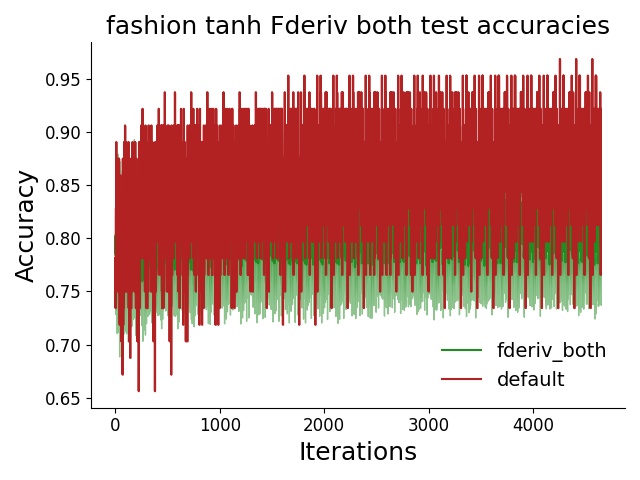} 
    \caption{Fashion both nonlinear derivative} 
  \end{subfigure}%%
  \begin{subfigure}[b]{0.5\linewidth}
    \centering
    \includegraphics[width=0.75\linewidth]{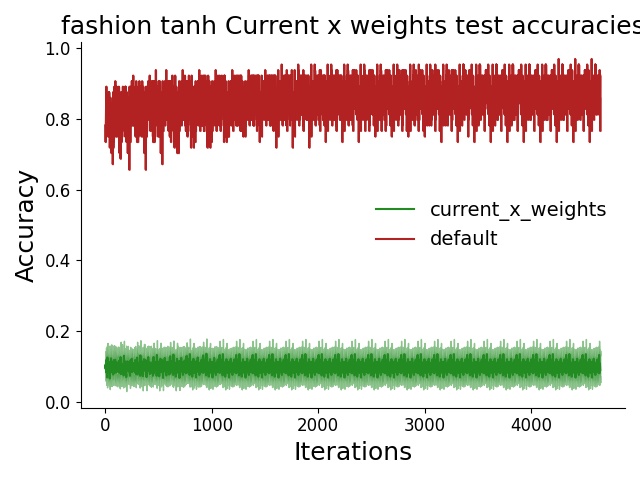} 
    \caption{Fashion current x weight update} 
  \end{subfigure} 
  \caption{Test accuracy for the FashionMNIST dataset through different relaxations of the frozen feedforward pass assumption. Similar to the MNIST results, we find that relaxing the assumption on both x update (Equation 1) and weight update (Equation 2) nonlinear derivatives does not affect performance, while relaxing the $\bar{x}$ term in the weight update equation destroys performance.}
\end{figure} 

\section*{Appendix C: CNN Architecture and Results}

For the CNN experiments, we used a simple CNN architecture which is designed simply to test these simplifications and not necessarily obtain state of the art performance. This was due to the greater computational cost of the AR algorithm compared to backprop since the AR algorithm contains a relaxation phase which requires multiple iterations which each have the same cost as backprop. Our CNN consisted of a convolutional layer followed by a max-pooling layer, followed by an additional convolutional layers, then two fully connected layers. The convolutional layers had 32 and 64 filters respectively, while the FC layers had 64,120,and 10 neurons respectively. For CIFAR100 there are 100 output classes so the final layer had 100 neurons. The labels were one-hot-encoded and fed to the network. All input images were normalized so that their pixel values lay in the range $[0,1]$ but no other preprocessing was undertaken. We used hyperbolic tangent activations functions at every layer except the final layer which was linear. The network was trained on a mean-square-error loss function. 

We tested the simplifications to AR proposed in the companion paper -- of removing the nonlinear derivatives and using learnable backwards weights on this CNN architecture on three more challenging datasets (CIFAR, CIFAR100, and SVHN). In the main text only the CIFAR results are presented. Here we present the SVHN and CIFAR100 results. We tested the simplifications (dropping nonlinearity or learning backwards weights) on just the convolutional layers of the network, just the fully-connected layers of the network, or both together. We found that ultimately performance was largely maintained even when both convolutional and fully connected layers in the network used learnable backwards weights or had their nonlinear derivatives dropped from both the update and weight equations. These results speak to the scalability and generalisability of the simplifications introduced in the companion paper and the general robustness of the AR algorithm. We implemented the learnable backwards weights of the CNN by applying Equation 3 to the flattened form of the CNN filter kernel weights.

 \begin{figure}[htb]
    \centering % <-- added
\begin{subfigure}{0.25\textwidth}
  \includegraphics[width=\linewidth]{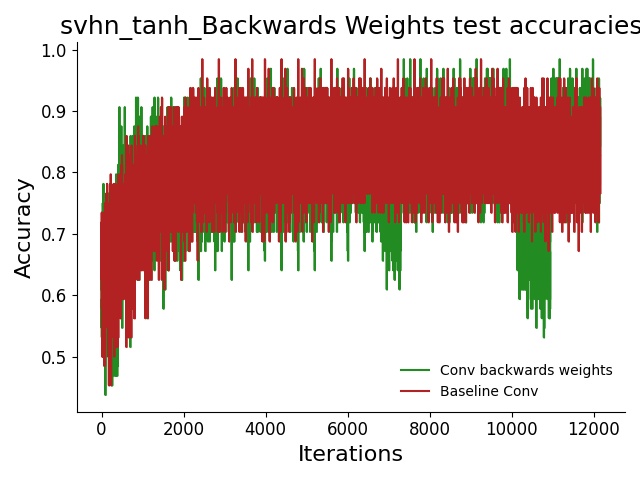}
  \caption{Conv backwards weights}
\end{subfigure}\hfil % <-- added
\begin{subfigure}{0.25\textwidth}
  \includegraphics[width=\linewidth]{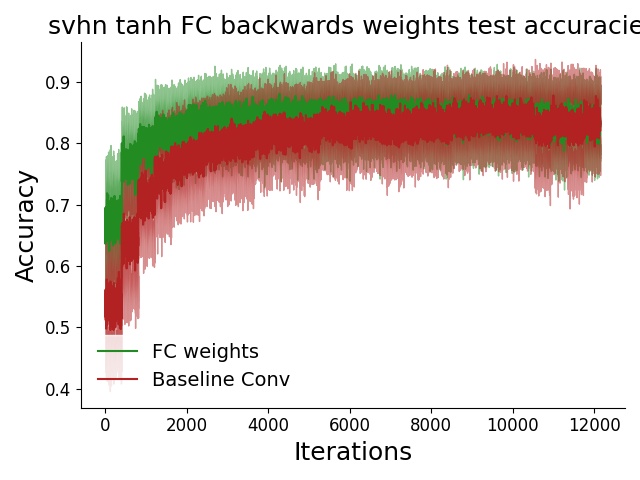}
  \caption{FC backwards weights}
\end{subfigure}\hfil % <-- added
\begin{subfigure}{0.25\textwidth}
  \includegraphics[width=\linewidth]{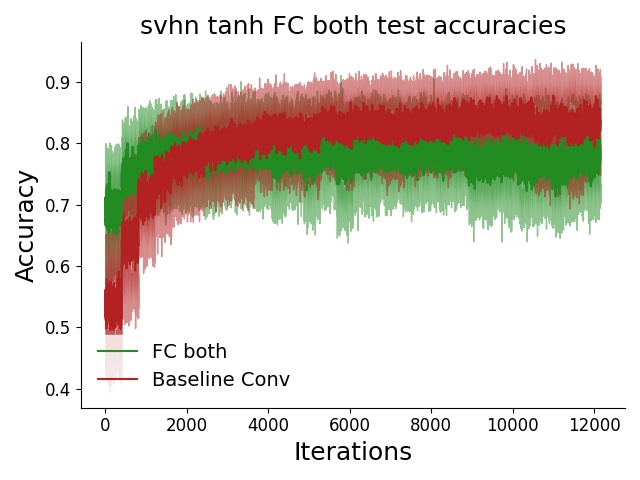}
  \caption{Both backwards weights}
\end{subfigure}

\medskip
\begin{subfigure}{0.25\textwidth}
  \includegraphics[width=\linewidth]{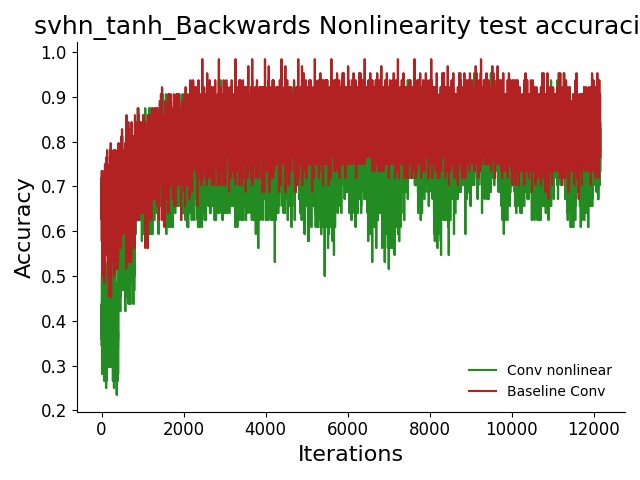}
  \caption{Conv no nonlinear derivative}
\end{subfigure}\hfil % <-- added
\begin{subfigure}{0.25\textwidth}
  \includegraphics[width=\linewidth]{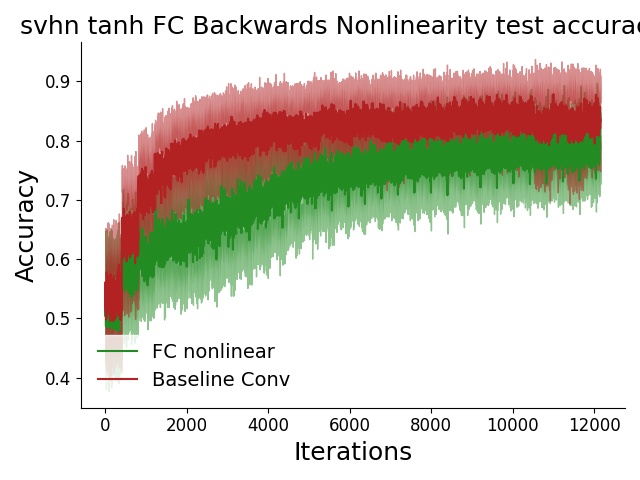}
  \caption{FC no nonlinear derivative}
\end{subfigure}\hfil % <-- added
\begin{subfigure}{0.25\textwidth}
  \includegraphics[width=\linewidth]{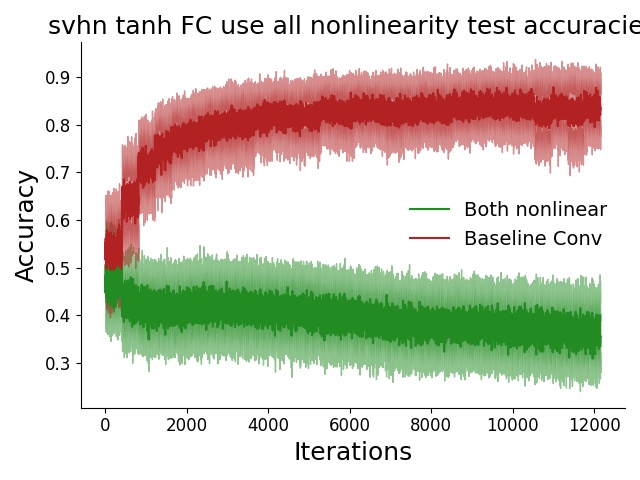}
  \caption{Both no nonlinear derivative}
\end{subfigure}
%\caption{}
\label{fig:images}
\end{figure}

We see that the simplifications also scale to the CNN for the SVHN dataset, although, interestingly, performance is degraded on this dataset when both convolutional and FC nonlinearities are dropped. However, since this does not occur in the other, more challenging, CIFAR datasets, we take this result to be an anomaly.

 \begin{figure}[htb]
    \centering % <-- added
\begin{subfigure}{0.25\textwidth}
  \includegraphics[width=\linewidth]{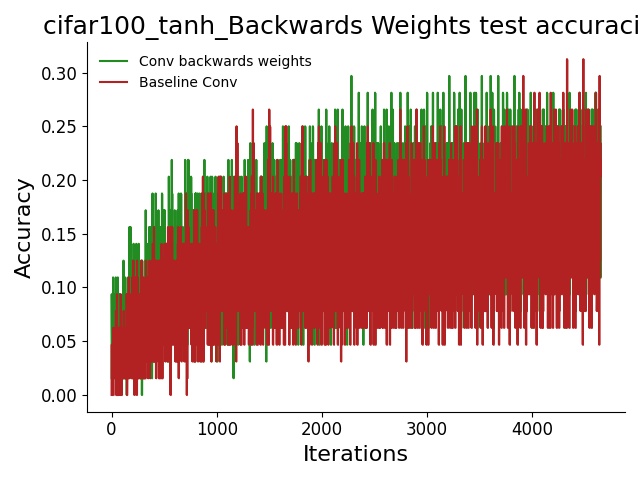}
  \caption{Conv backwards weights}
\end{subfigure}\hfil % <-- added
\begin{subfigure}{0.25\textwidth}
  \includegraphics[width=\linewidth]{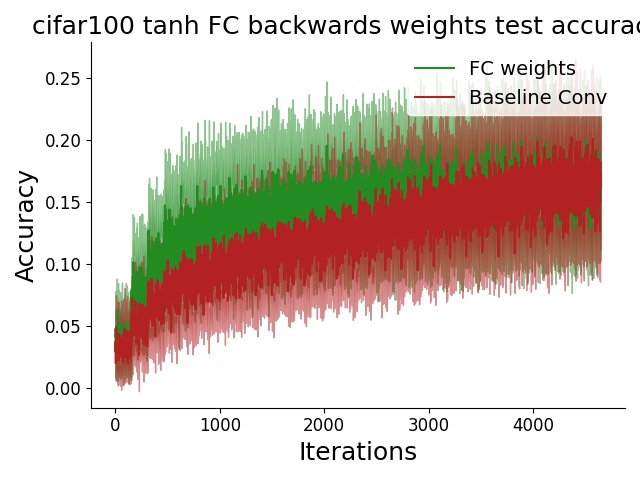}
  \caption{FC backwards weights}
\end{subfigure}\hfil % <-- added
\begin{subfigure}{0.25\textwidth}
  \includegraphics[width=\linewidth]{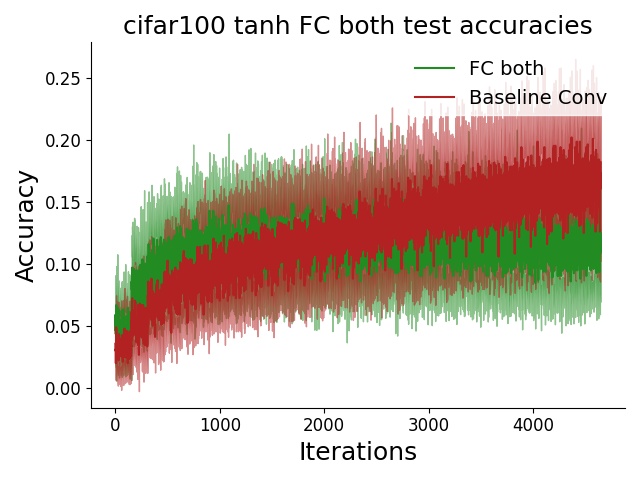}
  \caption{Both backwards weights}
\end{subfigure}

\medskip
\begin{subfigure}{0.25\textwidth}
  \includegraphics[width=\linewidth]{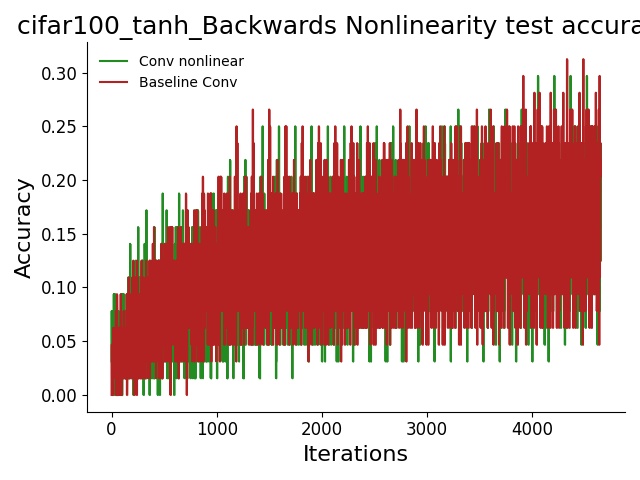}
  \caption{Conv no nonlinear derivative}
\end{subfigure}\hfil % <-- added
\begin{subfigure}{0.25\textwidth}
  \includegraphics[width=\linewidth]{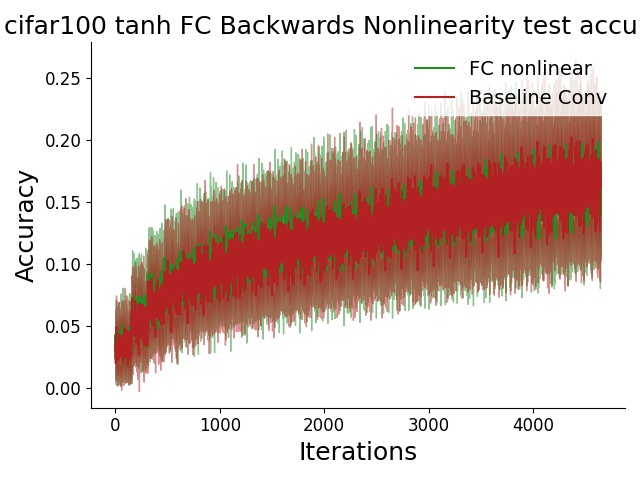}
  \caption{FC no nonlinear derivative}
\end{subfigure}\hfil % <-- added
\begin{subfigure}{0.25\textwidth}
  \includegraphics[width=\linewidth]{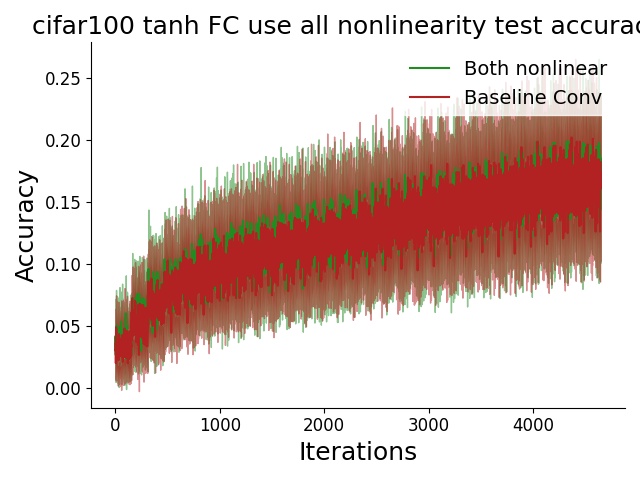}
  \caption{Both no nonlinear derivative}
\end{subfigure}
%\caption{}
\end{figure}

Here we see that the simplifications also scale comparably to the full AR algorithm (and thus backprop) in the most challenging CIFAR100 dataset.

\end{document}